\documentclass{article} 
\usepackage{sles,times}
\iclrfinalcopy

\usepackage{caption}
\usepackage{subcaption}
\usepackage{amsmath}
\usepackage{amsfonts}
\usepackage{bm}
\usepackage{hyperref}
\usepackage[capitalise]{cleveref}
\usepackage{url}
\usepackage{amsfonts}
\usepackage{amssymb}
\usepackage{scalerel}
\usepackage{multirow}
\usepackage{adjustbox}
\usepackage{algorithm}
\usepackage{algpseudocode}
\usepackage{booktabs}

\DeclareMathOperator*{\Fuse}{\scalerel*{\oplus}{\sum}}
\DeclareMathOperator*{\fuse}{\scalerel*{\oplus}{\textstyle\sum}}

\DeclareMathOperator*{\discount}{\scalerel*{\otimes}{\textstyle\sum}}
\newcommand{\up}{$\uparrow$}
\newcommand{\dn}{$\downarrow$}

\title{Subjective Logic Encodings}

\author{Jake Vasilakes\thanks{Work completed while a PhD student at the University of Manchester.} \\
Department of Computer Science\\
University of Sheffield \\
\texttt{j.vasilakes@sheffield.ac.uk}
\And
Chrysoula Zerva \\
Instituto de Telecomunicaçoes, ELLIS Unit Lisbon\\
Instituto Superior Tecnico, 
Universidade de Lisboa\\
\texttt{chrysoula.zerva@tecnico.ulisboa.pt}
\And
Sophia Ananiadou \\
National Centre for Text Mining,
Department of Computer Science, The University of Manchester \\
Artificial Intelligence Research Center,
National Institute of Advanced Industrial Science and Technology \\
\texttt{sophia.ananiadou@manchester.ac.uk}}

\begin{document}

\maketitle

\begin{abstract}
Many existing approaches for learning from labeled data assume the existence of gold-standard labels. According to these approaches, inter-annotator disagreement is seen as noise to be removed, either through refinement of annotation guidelines, label adjudication, or label filtering.
However, annotator disagreement can rarely be totally eradicated, especially on more subjective tasks such as sentiment analysis or hate
speech detection where disagreement is natural. Therefore, a new approach to learning from labeled data, called data perspectivism, seeks
to leverage inter-annotator disagreement to learn models that stay true to the inherent uncertainty of the task by treating annotations as opinions of the annotators, rather than gold-standard facts. Despite this conceptual grounding, existing methods under
data perspectivism are limited to using disagreement as the sole source of annotation uncertainty. To expand the possibilities of data perspectivism,
we introduce Subjective Logic Encodings (SLEs), a flexible framework for constructing classification targets that explicitly encodes annotations as
opinions of the annotators. Based on Subjective Logic Theory, SLEs encode labels as Dirichlet distributions and provide principled methods
for encoding and aggregating various types of annotation uncertainty —annotator confidence, reliability, and disagreement— into the targets.
We show that SLEs are a generalization of other types of label encodings as well as how to estimate models to predict SLEs using a distribution
matching objective. We make our code publicly available at \url{https://github.com/jvasilakes/SLEncodings}.
\end{abstract}

\section{Introduction}

Machine learning is inherently biased. \citet{Hovy_Prabhumoye_2021} identify five sources of bias: data selection, annotation, input representations, model, and experimental design. While all five have been studied in detail in previous works, a new view on annotation bias called \textit{data perspectivism} \citep{Basile_Cabitza_Campagner_Fell_2021} has gained traction. Data perspectivism re-frames annotations as opinions of the annotators rather than ``gold-standard'' facts. It also stands in contrast to existing methods for learning from noisy labels, which attempt to eradicate noise rather than embrace it \citep{Song_Kim_Park_Shin_Lee_2022}. This movement is inspired by the myth that there ought to be a single true label and that annotator disagreement is to be avoided \citep{Aroyo_Welty_2015}. Data perspectivism thus aims to embrace and leverage annotation uncertainty to build ML systems that are more representative of the event being modeled.

\subsection{Contributions}

Despite this conceptual grounding, existing methods under data perspectivism focus on utilizing only one of many types of annotation uncertainty ---annotator disagreement--- resulting incomplete opinion representations. To expand on this, we propose Subjective Logic Encodings (SLE), a flexible framework for constructing target distributions for classification tasks which explicitly encodes annotations as subjective opinions of the annotators. SLEs are based on Subjective Logic Theory \citep{Josang_2016}, which provides principled methods for encoding and aggregating various types of annotation uncertainty ---annotator disagreement, as well as reliability and subjective uncertainty---  into target Dirichlet distributions. We estimate a neural network to predict these distributions using a simple distribution matching objective, and show that this is equivalent to standard cross-entropy training with gold-standard labels in the absence of uncertainty. Experiments with both synthetic and real datasets in natural language processing and computer vision show that SLEs are more general and flexible method for learning from annotation uncertainty while matching the performance of existing methods.

\section{Sources of Uncertainty in Annotation}
\label{sec:uncertainty_sources}

We are interested in modeling \textit{aleatoric uncertainty}, also called data uncertainty, which refers to the irreducible uncertainty inherent in the data or events being modeled. It stands in contrast to \textit{epistemic uncertainty} ---due to a lack of knowledge regarding the events--- and \textit{distributional uncertainty} ---uncertainty regarding out-of-distribution (OOD) events. Generally, aleatoric uncertainty arises from noise inherent in the event being modeled (e.g., a dice roll) or from measurement noise \citep{Gal_2016}. More specifically, we view annotation as a measurement process and break it down into three indicators of uncertainty: inter-annotator disagreement indicates inherent event uncertainty, while annotator reliability and confidence indicate measurement uncertainty. We discuss each of these indicators below.

\paragraph{Annotator Reliability:}
An annotator may be highly confident in their annotations but still often provide incorrect or inconsistent labels due to a lack of expertise, the difficulty of the annotation task, etc. Conversely, adversarial annotators may purposefully provide poor annotations. Annotator reliability is thus an indicator of measurement noise. Simple measures of reliability are average agreement between annotators \citep{Tratz_Hovy_2010}, average Cohen's $\kappa$ between pairs of annotators \citep{Hovy_Berg-Kirkpatrick_Vaswani_Hovy_2013}, and the G-index \citep{Gwet_2008}. Previous works have also learned models of annotator  reliability directly from data \citep{Hovy_Berg-Kirkpatrick_Vaswani_Hovy_2013, Jagabathula_Subramanian_Venkataraman_2017, Li_Fahlstrom_Myrman_Mu_Ananiadou_2019}.

\paragraph{Annotator Confidence:}
Borderline examples, difficult annotation tasks, or unclear annotation guidelines can all influence an annotator's uncertainty in their assigned labels. Thus individual annotator confidence is an indicator of measurement noise. Previous works have measured confidence by asking annotators additional questions \citep{Nguyen_Valizadegan_Hauskrecht_2014} or eliciting probabilistic labels \citep{Collins_Bhatt_Weller_2022}.

\paragraph{Inter-Annotator Disagreement:}
Even assuming perfect annotator reliability and confidence, multiple annotators may still assign different labels to the same event, indicating an inherent uncertainty in the even \citep{Plank_Hovy_Sogaard_2014}. Inter-annotator disagreement is thus an indicator of aleatoric uncertainty. There has been a recent surge of interest in leveraging disagreements between annotators for improving classification (see \citet{Uma_Fornaciari_Hovy_Paun_Plank_Poesio_2021} for a survey).

It is commonly assumed that crowd annotations are samples from the same categorical distribution over labels for a given example (e.g., this assumption is explicitly stated in \cite{Baan_Aziz_Plank_Fernandez_2022}). The existence of annotator confidence and reliability, however, suggest that such samples pass through a intermediate ``measurement'' process, i.e., the subjective opinion of the annotator, which lends uncertainty to each annotation. It is therefore necessary to explicitly encode annotations as opinions ---i.e., including information regarding annotator confidence and reliability. In a more general sense, we ought to encode both first-order annotation uncertainty (i.e., probability of each label in a categorical distribution) as well as second-order uncertainty (i.e., uncertainty regarding the probabilities themselves). The next section discusses how to do this using Subjective Logic.

\section{Background: Subjective Logic Theory}
\label{sec:slt}

The above shows that it is natural to view annotations as subjective opinions of the annotators. However, the second-order uncertainty we aim to encode is incompatible with the current \textit{status quo} of vector representations, which are limited to first order uncertainty. Subjective Logic (SL) is a type of probabilistic logic that explicitly encodes events as the opinions of agents, with separate dimensions for first- and second-order uncertainty.
We here provide a brief overview of the aspects of SL necessary to understand SLEs.

\subsubsection{Opinion Representation}

In SL, a subjective opinion regarding some item $i$ according to an agent $m$ over a domain of $K$ possible events (e.g., class labels) is denoted $\omega^{(i)}_m = (\bm{b}^{(i)}_m, u^{(i)}_m, \bm{a})$, where $\bm{b}^{(i)}_m \in [0,1]^K$ is a vector of beliefs, $u^{(i)}_m \in [0,1]$ represents the uncertainty of the opinion, and $\bm{a} \in [0,1]^K$ is a vector of base rates or prior probabilities over the event space\footnote{Without loss of generality, we assume uniform priors throughout this work.}. These parameters are subject to the constraint
$u^{(i)}_m + \sum_{k=1}^K b^{(i)}_{m,k} = 1$.
SL opinions can be reparameterized as Dirichlet distributions, and the mapping to Dirichlet parameters $\bm{\alpha}$ is

\vspace*{-1em}
\begin{equation}
    \bm{\alpha}^{(i)}_m = \frac{2\bm{b}^{(i)}_m}{u^{(i)}_m} + K\bm{a}
\end{equation}

The expectation of this Dirichlet can be computed from the SL opinion parameters.

\begin{equation}\label{eq:prob}
    \bm{P}^{(i)}_m = \mathbb{E}[\omega^{(i)}_m] = \bm{b}^{(i)}_m + u^{(i)}_m \bm{a}
\end{equation}

\subsubsection{Combining Opinions}

A key feature of SL opinions is that they may be combined to form consensus opinions using various operators. In this work, we utilize the cumulative belief fusion and trust discounting operators. Cumulative belief fusion, denoted $\fuse$, combines two opinions regarding a single event ---such as two annotators $m$ and $q$ observing a single example--- treating each as evidence of the true label distribution. It is defined as

\begin{equation}\label{eq:cumulative_fusion}
    \omega_{[m\lozenge q]} = \omega_m \oplus \omega_q =
    \begin{cases}
        \bm{b}_{[m\lozenge q]} = \frac{\bm{b}_m u_q + \bm{b}_q u_m}{u_m + u_q - u_m u_q} \\[5pt]
        u_{[m\lozenge q]} = \frac{u_m u_q}{u_m + u_q - u_m u_q} \\[5pt]
        \bm{a}_{[m \lozenge q]} = \frac{\bm{a}_m u_q + \bm{a}_q u_m - (\bm{a}_m + \bm{a}_q) u_m u_q}{u_m + u_q - 2 u_m u_q}
    \end{cases}
\end{equation}

Cumulative fusion reduces the uncertainty $u$ by combining evidence. This means that as more opinions are fused, the uncertainty tends towards zero and the resulting Dirichlet approaches zero variance, equivalent to a categorical probability.

The trust discounting operator increases the uncertainty of an opinion according to a separate opinion of the reliability of that annotator. It is denoted $\discount$ and defined as 

\begin{equation}
    \hat{\omega}^{(i)}_m = \omega^{(m)}_R \discount \omega^{(i)}_m = 
    \begin{cases}
        \hat{\bm{b}}^{(i)}_m & = \bm{P}^{(m)}_R \bm{b}^{(i)}_m \\
        \hat{u}^{(i)}_m & = 1 - \bm{P}^{(m)}_R ~\sum^K_{k=1} b^{(i)}_{m,k} \\
        \hat{\bm{a}}_m       & = \bm{a}_m \\
    \end{cases}
\end{equation}

\noindent
where $\omega^{(m)}_R$ is an opinion of the reliability of the subjective opinion $\omega^{(i)}_m$.

\section{Constructing SLEs}
\label{sec:sle}

We can utilize cumulative fusion and trust discounting to encode and aggregate labels as SLEs. Let there be a dataset of $N$ annotated examples $\mathcal{D} = \{\bm{x}^{(i)}, \bm{y}^{(i)}\}_{i=1}^N \in (\mathcal{X}, \mathcal{Y})^N$, where $\mathcal{X} \in \mathbb{R}^d$ is the input space and $\mathcal{Y} \in \{1,...,K\}^M$ is the label space over $K$ labels and $M$ annotators. That is, each $\bm{y}^{(i)}$ is a vector of class labels from all $M$ annotators\footnote{For simplicity, we assume each example has the same number of annotators, but we note that this is not a requirement for label aggregation.}.
Further, each individual judgment $y^{(i)}_m$ has metadata regarding the annotator's reliability $r^{(i)}_m$ and subjective uncertainty $u^{(i)}_m$\footnote{Our approach is agnostic to whether reliability is an overall score for the annotator \cite{Hovy_Berg-Kirkpatrick_Vaswani_Hovy_2013} or an instance level score from another model \cite{Li_Fahlstrom_Myrman_Mu_Ananiadou_2019}}. Given this, our goal for constructing the target distributions is twofold:

\begin{enumerate}
    \item Encode each individual judgement taking into account each of the sources of uncertainty described in \cref{sec:uncertainty_sources}.
    \item Define a method for aggregating individual encoded judgments into a target distribution for training a classification model that accounts for label disagreement.
\end{enumerate}

As discussed in \cref{sec:slt}, the opinion of annotator $m$ regarding the $i^{th}$ example is $\omega^{(i)}_m = (\bm{b}^{(i)}_m, u^{(i)}_m, \bm{a}_m)$. In practice, $u^{(i)}_m$ can be any mapping from a user-supplied indicator of subjective uncertainty to the range $[0,1]$. The belief vector $\bm{b}^{(i)}_m$ is computed using the subjective uncertainty as

\begin{equation}
    \bm{b}^{(i)}_m = \tilde{\bm{y}}^{(i)}_m - u^{(i)}_m \tilde{\bm{y}}^{(i)}_m
\end{equation}

where $\tilde{\bm{y}}^{(i)}_m$ is the vector encoding of $y^{(i)}_m$, such as one-hot or probabilities\footnote{For binary labels ($K=2$), the distribution is a Beta with $\alpha$ or $\beta = 1$, equivalent to a Kumaraswamy distribution \cite{Kumaraswamy_1980}.}.

Finally, we compute the SLE as the aggregate of the opinions for a given example $x$ using the cumulative fusion \cref{eq:cumulative_fusion} and trust discounting operators \cref{eq:trust_discount}

\begin{equation}\label{eq:trust_discount}
    \omega^{(i)}_{\lozenge} = \Fuse^M_{m=1} \omega^{(m)}_R \discount \omega^{(i)}_m
\end{equation}

In the case where there is no information regarding reliability ($r^{(i)}_m = 1$) or uncertainty ($u^{(i)}_m = 0$), this process results in a dogmatic opinion $\omega^{(i)}_{\lozenge} = (\bm{b}^{(i)}_{\lozenge}, 0, \bm{1}/K)$ which is equivalent to a categorical probability. In the further case where there is no disagreement between annotators, the resulting opinion is equivalent to a one-hot encoded target.


\section{Synthetic Data Experiments}
\label{sec:synth_data}

To our knowledge, there are no real-world datasets that contain all the sources of uncertainty discussed in this work. We therefore illustrate the benefit of using SLEs in place of other methods for label aggregation using synthetic data. We generate synthetic datasets to assess the ability of SLEs to recover the true label from a set of crowd annotations that have been corrupted according to a range of annotator reliabilities and confidences. We conduct three experiments: \textbf{(1)} assuming perfect certainty, we vary total annotator reliability from very low to perfect; \textbf{(2)} assuming high reliability, we vary annotator confidence from very low to perfect; \textbf{(3)} as for (2), but we assume low annotator reliability.

\subsection{Data Generation}
We generate a set of $N$ 5-class true labels as evenly spaced points on the simplex. For each true label, we generate crowd annotations from $M=10$ annotators
with varying degrees of confidence and reliability. Specifically, we draw values of these parameters from Beta distributions and use them to corrupt the true labels: a true label is permuted with probability equal to reliability and it is recalibrated according to confidence. This process is detailed in \cref{alg:gen_proc} below, where $\text{permute}(\bm{y}, r)$ shuffles the indices of $\bm{y}$ with probability $r$ and $\text{recalibrate}(\bm{y}, c) = \frac{\text{exp}(\ln \bm{y}_i c)}{\sum^d_{j=1}\text{exp}(\ln \bm{y}_j c)}$ smooths the probabilities in $\bm{y}$ according to the confidence $c$. 
This process is an identity when $c = 1$ and $r = 1$ and pushes $\bm{y}$ towards the uniform distribution as $c \rightarrow 0$. Annotator certainty and reliability can thus be varied by specifying different $\alpha$ and $\beta$ parameters to the beta distributions.

\begin{algorithm}
\caption{Synthetic Data Generation Process}\label{alg:gen_proc}
\begin{algorithmic}[1]
    \Require Two sets of Beta distribution parameters $(\alpha, \beta)$ and $(\alpha', \beta')$
    \For{$m \in M$}
        \State $c \sim Beta(\alpha, \beta)$ \Comment{Sample annotator confidence.}
        \State $r \sim Beta(\alpha', \beta')$ \Comment{Sample annotator reliability.}
        \For{$\bm{y} \in \triangle^{(d-1)}$}
            \State $\bm{y}^* \gets \text{permute}(\bm{y}, r)$ \Comment{Permute $\bm{y}$ with probability $r$.}
            \State $\bm{y}^* \gets \text{recalibrate}(\bm{y}^*, c)$ \Comment{Recalibrate probabilities.}
        \EndFor
    \EndFor
\end{algorithmic}
\end{algorithm}

Using this process we generate synthetic datasets in three different scenarios according to different values of the $\alpha$ and $\beta$ parameters. In the first, we assume no information regarding annotator confidence ($c=1$) and only vary the reliability of the annotators (\cref{fig:reliability}).
In the second, we assume high reliability and vary annotator confidence (\cref{fig:conf_high_rel}). In the third, we assume low reliability and vary annotator confidence (\cref{fig:conf_low_rel}). We report all $\alpha$ and $\beta$ parameters used in \cref{app:hyperparams}.

\subsection{Evaluation}

\newcommand{\argmax}{\operatornamewithlimits{argmax}}
The evaluation uses both hard and soft evaluation metrics between the aggregated crowd annotations and the true labels. For the hard metric we report the micro-averaged F1 on the test set, and label predictions are obtained from SLEs by taking the argmax of the mode of the predicted Dirichlet, i.e., $\hat{y} = \argmax_k \mathbb{E}_{Q_{\theta}}(y_k|\bm{x})$. Where the F1 measures the ability of the model to make correct predictions, the soft evaluation metrics measure how well the model is able to match the distribution of the crowd annotations. Specifically, we use the Jensen-Shannon Divergence (JSD) \citep{Lin_1991} and the Normalized Entropy Similarity (NES) \citep{Uma_Fornaciari_Hovy_Paun_Plank_Poesio_2021} between the aggregated and true distributions. 

For each experiment, we plot the F1, JSD, and NES curves over the range of uncertainty values. We compare SLEs to two baseline aggregation methods: Majority Voting (MV) chooses the class with the greatest frequency in the crowd annotations, resulting in a one-hot label; Soft Voting (Soft) computes the average count of each label from the crowd annotations, resulting in a vector of continuous labels. We evaluate each aggregation method using all crowd annotations as well as a subset filtered according to reliability as in \cite{Tratz_Hovy_2010}. For all experiments, we average the results over 10 runs with crowd-annotations generated according to different random seeds.

\subsection{Results}

\begin{figure}[t]
    \centering
    \begin{subfigure}{\textwidth}
        \centering
        \includegraphics[width=\textwidth]{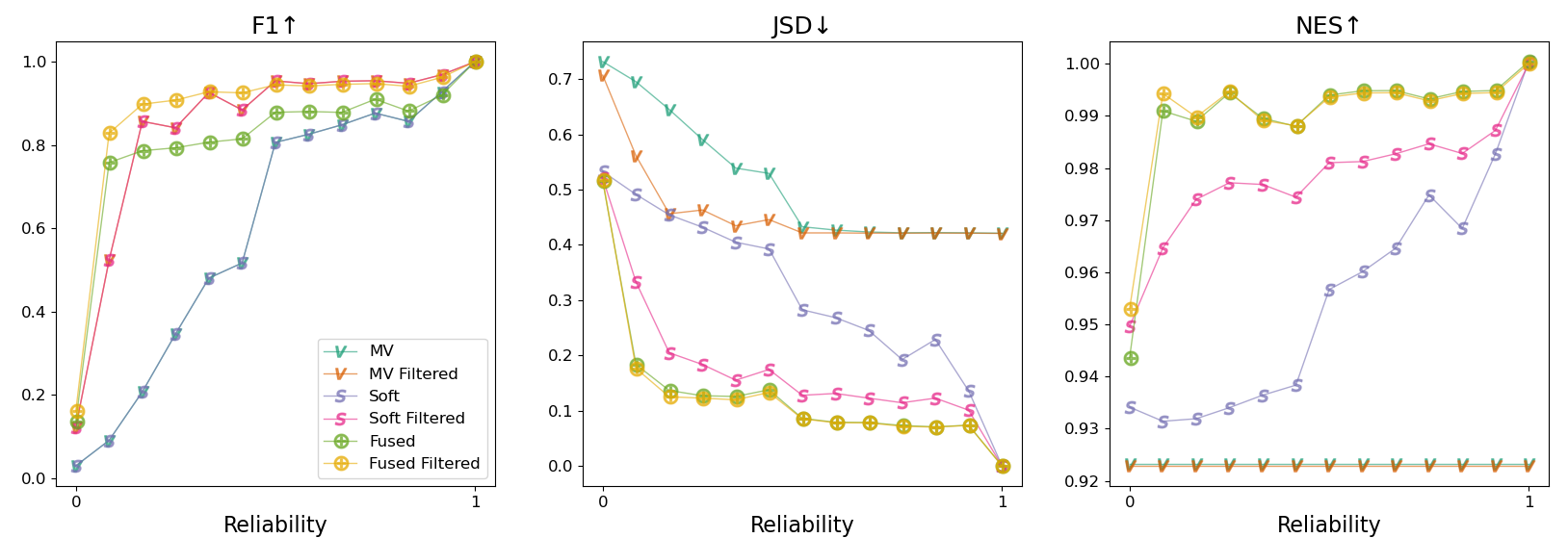}
        \caption{Varying reliabilities from very low $r \sim Beta(1, 10)$ to perfect $r=1$, assuming perfect confidence $c=1$ for all annotators.}
        \label{fig:reliability}
    \end{subfigure}
    \hfill
    \begin{subfigure}{\textwidth}
        \centering
        \includegraphics[width=\textwidth]{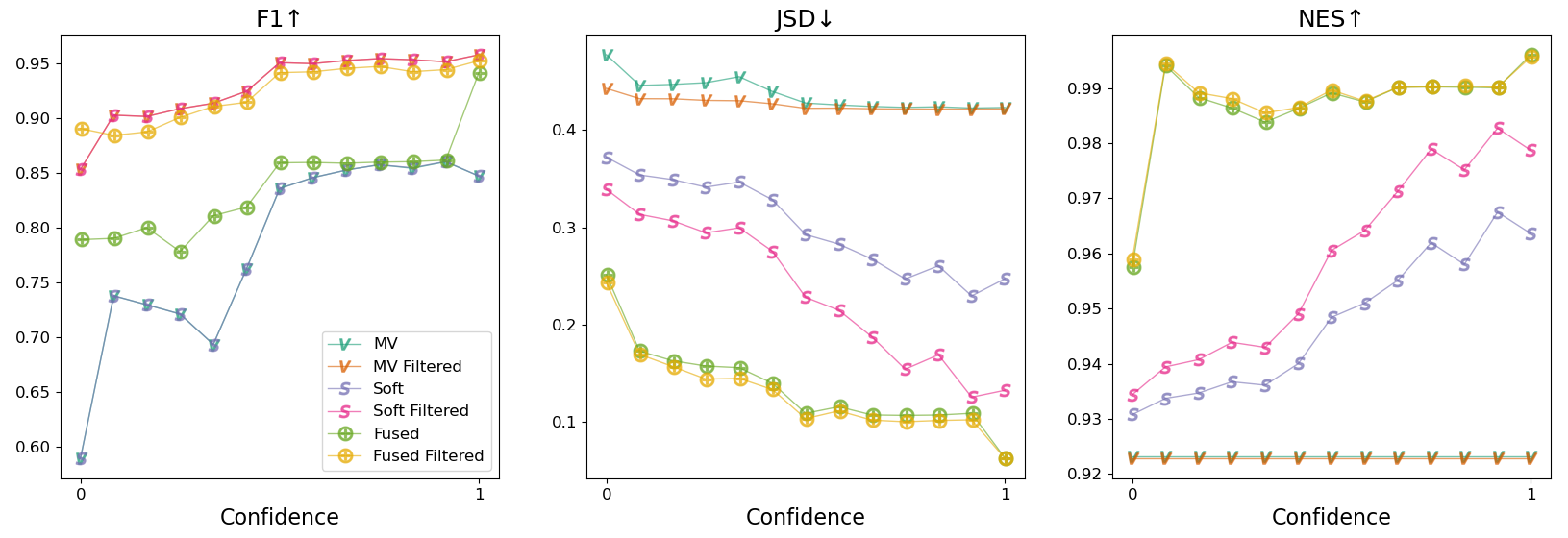}
        \caption{Varying confidences from very low $c \sim Beta(1, 10)$ to perfect $c=1$, assuming high reliability $r \sim Beta(10, 1)$.}
        \label{fig:conf_high_rel}
    \end{subfigure}
    \hfill
    \begin{subfigure}{\textwidth}
        \centering
        \includegraphics[width=\textwidth]{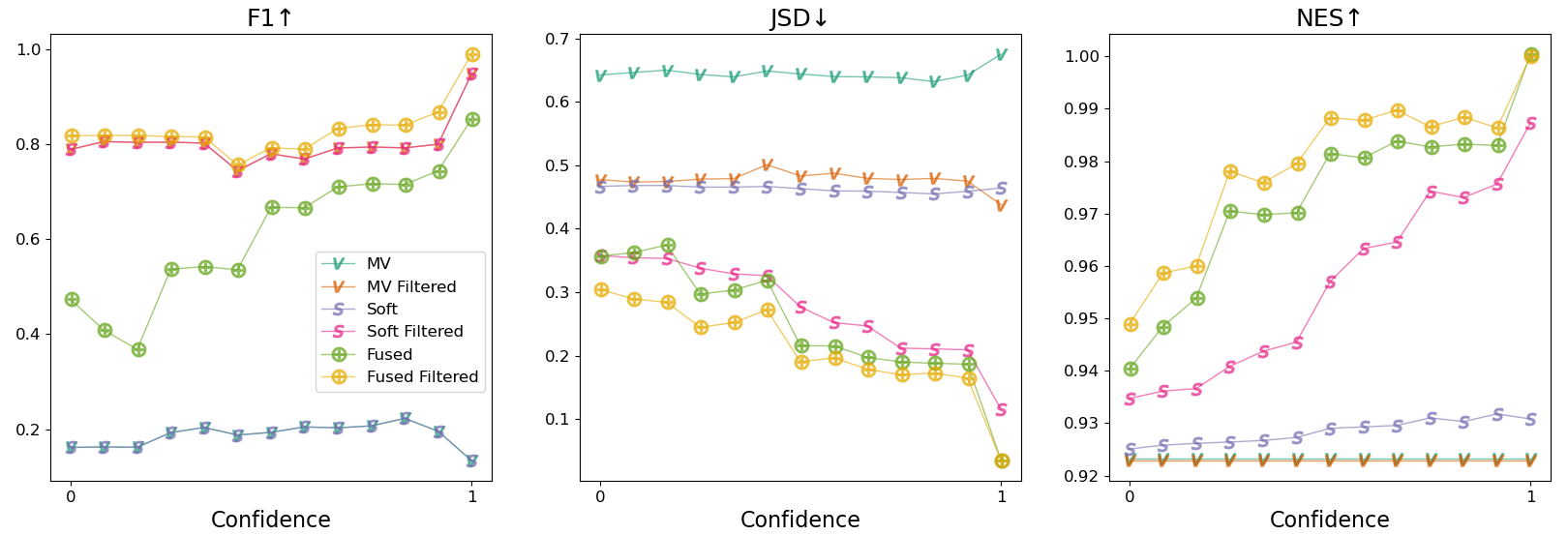}
        \caption{Varying confidences from very low $c \sim Beta(1, 10)$ to perfect $c=1$, assuming low reliability $r \sim Beta(1, 10)$.}
        \label{fig:conf_low_rel}
    \end{subfigure}
    \caption{F1, JSD, and NES results for Majority Voting (MV), Soft voting (Soft), and SLE cumulative fusion (Fused) given both all and filtered annotations on the synthetic data across different ranges of uncertainty. Results using CrowdTruth were nearly identical to Soft, and so omitted here for clarity.}
    \label{fig:synth_results}
\end{figure}

The plots in \cref{fig:synth_results} show that SLEs are better able to recover the true distribution from crowd annotations than the baselines, measured by JSD and NES. Additionally, the ability of SLEs to ``deweight'' annotations according to uncertainty means that they achieve higher F1s than the unfiltered baselines. Filtering annotations according to an estimate of annotator reliability improves results for all methods, but we note that the difference in JSD and NES between unfiltered and filtered SLEs is much less than MV and Soft, suggesting the SLEs are better able to capture the true distribution even in the presence of noisy labels.

\section{Learning SLEs}

The sources of aleatoric uncertainty described in \cref{sec:uncertainty_sources} have been independently studied in previous works. However, the methods used for incorporating them into classification models are quite specialized and generally incompatible with each other. In this section, we describe in detail how SLEs aggregate each of these sources of uncertainty into a single target for model estimation.

\subsection{Dirichlet Neural Networks}
\label{sec:dirichlet_nns}

Using SLEs in machine learning applications requires a classifier to predict Dirichlet distributions. Predicting Dirichlets has been studied extensively in previous work. \cite{Malinin_Gales_2018} introduce Prior Networks, which use Dirichlet outputs to model distributional uncertainty given out-of-distribution data. Concurrently, and most similar to our work, \cite{Sensoy_Kaplan_Kandemir_2018} propose a Dirichlet neural network for modeling prediction uncertainty that is explicitly influenced by SL. Still, this work focuses on inducing predictive uncertainty of the model given gold-standard labels rather than utilizing aleatoric uncertainty obtained from data. Also, despite being inspired by SL, their model predicts the $\bm{\alpha}$ parameters of a Dirichlet instead of the corresponding SL opinion parameters, as we propose to do here. \cite{Joo_Chung_Seo_2020} develop a Dirichlet latent variable model, trained using variational inference, to construct a prior over the predicted class probabilities given gold labels. In contrast with these previous works, which model distributional uncertainty given gold-standard labels, this work focuses on encoding aleatoric uncertainty obtained as part of the annotation process.

\subsection{Model Estimation}

As described in \cref{sec:sle}, we assume a dataset with aggregated SLE target distributions $\mathcal{D} = \{\bm{x}^{(i)}, \omega^{(i)}_{\lozenge}\}_{i=1}^N \in (\mathcal{X}, \Omega)^N$. We aim to estimate a classification model $\mathcal{F}_{\Theta} : \mathcal{X} \rightarrow \Omega$ parameterized by $\Theta$ which maps from the input space $\mathcal{X}$ to the space of SL opinions $\Omega$ represented by reparameterized Dirichlet distributions. The process of estimating such a classifier amounts to distribution matching between the target $\omega^{(i)}_{\lozenge}$ and approximating SLE opinions $\omega^{(i)}_{\theta}$, which are predicted by a classifier $f_{\theta} \in \mathcal{F}_{\Theta}$. In our setup, a neural network $f_{\theta}$ predicts the $\bm{b}$ and $u$ parameters of the opinion representation.

\begin{equation}
    \bm{b}^{(i)}_{\theta}, u^{(i)}_{\theta} \leftarrow \text{softmax}(f_{\theta}(x^{(i)}))
\end{equation}

where softmax enforces the constraint $u^{(i)}_{\theta} + \sum_{k=1}^{K} b_{\theta,k}^{(i)} = 1$.

In the usual classification setup where the targets are gold-standard labels, model estimation often uses the cross entropy loss between the one-hot target labels $\tilde{\bm{y}}$ and the label probabilities predicted by the classifier, $Q_{\theta}$.

\begin{equation}\label{eq:ce_log}
    \mathcal{L}_{\theta} = \sum_{k=1}^K \tilde{y}_k ~\text{log}~ Q_{\theta}(\hat{y}=k|x)
\end{equation}

However, given that we are here concerned with target \textit{distributions} rather than labels, it may be clearer to rewrite the cross-entropy loss according to the KL divergence between the target distribution $P$ and predicted distribution $Q$.

\begin{equation}\label{eq:ce_kl}
    \mathcal{L}_{\theta} = D_{KL}(P~||~Q_{\theta}) + H(P)
\end{equation}

Since the entropy term in \cref{eq:ce_kl} is a constant that depends only on the target distribution, the minimization of the cross-entropy amounts to the minimization of the KL divergence between the target and the predicted distributions\footnote{In standard classification setups with a single ``gold'' judgment per sample, the target is one-hot, i.e., a deterministic categorical distribution over the $K$ labels. In this case, the the entropy of the target distribution $H(P)$ is zero and cross-entropy equals the KL-divergence. However, when using soft labels as in \cite{Peterson_Battleday_Griffiths_Russakovsky_2019,Uma_Fornaciari_Hovy_Paun_Plank_Poesio_2020}, the entropy term is $> 0$, meaning that the values of the cross-entropy loss between hard- and soft-label setups are not comparable.}. Thus our target loss function is simply the KL divergence, 

\begin{equation}\label{eq:kl_loss}
    \mathcal{L}_{\theta} = D_{KL}(P~||~Q_{\theta}) = \int_x P(x)~\text{log} \frac{P(x)}{Q_{\theta}(x)}
\end{equation}

which is essentially the same loss function proposed by \cite{Malinin_Gales_2018}, without the additional KL term for out-of-distribution data\footnote{It would be trivial to add support for OOD data using the additional KL term and OOD training data, but this is out of scope for this work.}. 

Unfortunately, there are two issues with this objective: (1) the KL divergence error surface is poorly suited to optimization when the target distributions are sparse ``one-hot'' distributions corresponding to dogmatic opinions; (2) this ``forward'' KL divergence penalizes $Q_{\theta}$ much less where $P$ is very small, which results in a $Q_{\theta}$ that covers a majority of the space and is a poor predictor.

To overcome (1), we follow \cite{Malinin_Gales_2018} and smooth dogmatic target opinions by redistributing a small $\epsilon$ amount of belief mass to the uncertainty parameter, which adds a small amount of density to the other corners of the probability simplex (\cref{eq:smoothing}).

\begin{equation}\label{eq:smoothing}
    \omega^{(i)}_m = 
    \begin{cases}
        \bm{b}^{(i)}_m & = \bm{b}^{(i)}_m - \epsilon \\
        u^{(i)}_m      & = u^{(i)}_m + \epsilon \\
        \bm{a}_m       & = \bm{a}_m
    \end{cases}
\end{equation}

To overcome (2), we follow \cite{Malinin_Gales_2019} and use the \textit{reverse} KL divergence

\begin{equation}\label{eq:rev_kl_loss}
    \mathcal{L}_{\theta} = D_{KL}(Q~||~P_{\theta}) = \int_x Q_{\theta}(x)~\text{log} \frac{Q_{\theta}(x)}{P(x)}
\end{equation}

which encourages $Q_{\theta}$ to fit under $P$. 

\section{Real-World Data Experiments}

Despite the lack of real-world datasets that contain all three sources of uncertainty, we can still compare SLEs to existing methods for learning from crowd annotations.
We experiment with the following real-world datasets in computer vision and natural language processing that contain crowd-sourced annotations with multiple annotators per example.

\begin{itemize}
    \item CIFAR-10S \citep{Collins_Bhatt_Weller_2022}: Image classification. A variant of the CIFAR-10 computer vision benchmark dataset \citep{hinton2012improving} annotated by 6 annotators per example. Additionally, CIFAR-10S elicited probabilistic labels from each annotator to capture each annotator's confidence. 
    \item MRE \citep{Dumitrache_Aroyo_Welty_2018}: A dataset of 4,000 sentences in English from PubMed abstracts expressing a \textit{cause} or \textit{treats} relation between two entities. Each example is annotated by 15 annotators.
    \item ConvAbuse \citep{CercasCurry_Abercrombie_Rieser_2021}: A collection of 7,000 sentences sampled from different chatbot systems annotated by 8 annotators for a variety of abuse types.
    \item RTE \citep{Hovy_Berg-Kirkpatrick_Vaswani_Hovy_2013}: 800 sentences pairs annotated for semantic entailment by 10 annotators. 
\end{itemize}

\subsection{Models}

We utilize the following models for each real-world dataset.

\begin{itemize}
    \item CIFAR-10S: Following previous work \citep{Collins_Bhatt_Weller_2022, Uma_Fornaciari_Hovy_Paun_Plank_Poesio_2020}, we employ a ResNet-32A model \citep{He_Zhang_Ren_Sun_2016} and replicate their evaluation setup.
    \item MRE: We use the same model and training procedure as \citet{Uma_Fornaciari_Hovy_Paun_Plank_Poesio_2021}. This is a fine-tuned BERT sentence classifier with a single linear output layer \citep{Devlin_Chang_Lee_Toutanova_2019}. 
    \item ConvAbuse: As for MRE, we use a fine-tuned BERT sentence classifier with a single linear output layer as in \citet{CercasCurry_Abercrombie_Rieser_2021}.
    \item RTE: We employ the model implemented by \citet{Uma_Fornaciari_Hovy_Paun_Plank_Poesio_2021}: The premise and hypothesis are concatenated, encoded by BERT, passed through 3 ReLU-activated linear layers, and a final linear output layer.
\end{itemize}

\subsection{Evaluation}

We compare SLEs to a number of baseline methods previously used to handle annotation uncertainty.

\begin{itemize}
    \item Majority-Voting (MV): the signal is the most common label from the annotator judgements. If more than one label is tied for most common, one of the them is chosen at random. Model estimation uses the cross-entropy loss.
    \item Cross-Entropy (CE) loss between the predicted distribution and the categorical distribution computed directly from the crowd labels \cite{Peterson_Battleday_Griffiths_Russakovsky_2019}.
    \item Kullback-Leibler divergence (KL) between the predicted distribution and the categorical distribution computed directly from the crowd labels \cite{Uma_Fornaciari_Hovy_Paun_Plank_Poesio_2021}.
    \item CrowdTruth: a method for aggregating labels that weights them according to measures of annotation, annotator, and example reliability \cite{Dumitrache_Inel_Aroyo_Timmermans_Welty_2018}. This method is similar to SLEs in what is measured, but is more prescriptive in how these measurements are computed. In fact, the measurements defined by CrowdTruth could be used by SLEs as measures of annotator uncertainty and reliability. Model estimation uses cross-entropy loss.
    \item Multi-task Learning from Soft Labels (MTLSL): a simple combination of learning from gold labels and categorical probabilities computed from the crowd labels \cite{Fornaciari_Uma_Paun_Plank_Hovy_Poesio_2021}. Model estimation uses cross-entropy for the gold labels and the reverse KL divergence for the probabilistic labels, as we do for SLEs.
\end{itemize}

A comparison of these methods to SLEs is given in table \cref{tab:method_comparison}. We note that, of the methods listed above, only CE and KL are able to learn from annotator confidence, but only via probabilistic labels, and only CrowdTruth is able to learn from annotator reliability. It is possible to combine MTLSL with CE/KL by eliciting soft labels from $>1$ annotator, but this still lacks the ability to learn from annotator reliability. SLEs, on the other hand, are able to incorporate all three sources of annotation uncertainty and do so without large changes to the model architecture. Furthermore, unlike MTLSL, SLEs do not require each example to be annotated by the same number of annotators, nor do they depend on having both gold-standard and crowd annotations available.

\newcommand{\chk}{\checkmark}
\begin{table}[h]
    \centering
    \begin{tabular}{lccccc|c}
        \toprule
                              &  MV  & CE   & KL   & CrowdTruth &  MTLSL & SLE   \\
        Annotator Confidence  &      & \chk & \chk &            &        & \chk  \\
        Annotator Reliability &      &      &      &  \chk      &        & \chk  \\ 
        Crowd Annotations     & \chk & \chk & \chk &  \chk      &  \chk  & \chk  \\
        \bottomrule
    \end{tabular}
    \caption{A comparison of methods for learning from annotation uncertainties.}
    \label{tab:method_comparison}
\end{table}

\subsection{Results}

We report the performance of SLEs compared to the baselines on the real world datasets in \cref{tab:real_results}. For all datasets besides ConvAbuse, which has only crowd annotations, we report results of methods using both gold and crowd annotations. 

\begin{table}[h]
    \centering
    \footnotesize
    \begin{adjustbox}{center}
    \begin{tabular}{lccc|ccc|ccc|ccc}
        \toprule
                   & \multicolumn{3}{c|}{CIFAR-10S}   & \multicolumn{3}{c|}{MRE}        & \multicolumn{3}{c|}{ConvAbuse}  &  \multicolumn{3}{c}{RTE}              \\
                   &  F1\up    &  JSD\dn  & NES\up    & F1\up    &  JSD\dn  & NES\up    &  F1\up   & JSD\dn   & NES\up    &   F1\up   &   JSD\dn   & NES\up       \\
        \midrule
        Gold       &  0.645    &  0.331   &{\bf0.390} &{\bf0.840}& 0.202    & 0.686     &          &          &           &  0.613    &  0.346     & 0.569        \\
        SLE (gold) &{\bf0.653} &{\bf0.326}&  0.369    & 0.830    & 0.198    &{\bf0.697} &   -      &  -       &  -        & {\bf0.617}&  0.351     &{\bf0.587}    \\
        MTLSL*     &  0.636    &  0.344   &  0.347    & 0.828    &{\bf0.145}& 0.596     &          &          &           &  0.610    & {\bf0.338} & 0.565        \\
        \midrule
        MV         &  0.645    &{\bf0.336}&  0.398    & 0.749    & 0.133    & 0.660     &  0.886   & 0.084    & 0.153     &  0.605    &  0.346     & 0.571        \\
        CE         &  0.661    &  0.356   &{\bf0.551} & 0.751    & 0.127    & 0.749     &  0.898   &{\bf0.072}& 0.375     & {\bf0.611}&  0.344     & 0.578        \\
        KL         &{\bf0.660} &  0.358   &{\bf0.551} & 0.751    & 0.127    & 0.749     &{\bf0.902}& 0.084    &{\bf0.414} &  0.609    & {\bf0.343} & 0.579        \\
        CrowdTruth &  0.633    &  0.643   &  0.547    & 0.756    & 0.122    & 0.750     &  0.897   & 0.315    & 0.336     &  0.603    &  0.354     & 0.589        \\
        SLE (crowd)&  0.649    &  0.364   &  0.546    &{\bf0.757}&{\bf0.120}&{\bf0.764} &  0.894   & 0.078    & 0.381     &  0.588    &  0.361     &{\bf0.592}    \\
        \bottomrule
    \end{tabular}
    \end{adjustbox}
    \caption{Hard and soft metric results on the real-world data sets. CIFAR-10S is the only real-world dataset that provides both annotator confidence and inter-annotator disagreement. ConvAbuse does not contain gold-standard labels. * MTLSL is trained using both gold and crowd annotations.}
    \label{tab:real_results}
\end{table}

\section{Discussion}

Results of the synthetic data experiments \cref{fig:synth_results} show a clear benefit of using SLEs over majority and soft voting for label aggregation, especially when information regarding annotator confidence and reliability are available. Results on the real-world datasets \cref{tab:real_results} are, however, mixed. SLEs with gold labels generally outperform standard cross-entropy training on gold labels as well as MTLSL, but this improvement may be an artifact of the different objective function. When using crowd annotations, CE and KL tend to outperform all other methods, although SLEs do perform best on the MRE dataset. As already discussed, a major limitation of these experiments is that none of the datasets contain all three sources of uncertainty. This means that even though SLEs are capable of modeling second-order uncertainty, they are not evaluated as such, since the targets contain only the types of uncertainty present in the datasets. 

Future research should provide a more in-depth evaluation of the learned SLEs, such as their interpretability and ability to use post-hoc measures of reliability and annotator confidence. For example, we might compute a simple estimate of annotator reliability as each annotator's average percentage agreement with the others over all training examples, or a measure of confidence by computing the normalized entropy of the soft labels obtained from each annotator in CIFAR-10S.

\section{Conclusion}

This paper introduced Subjective Logic Encodings (SLEs) a novel method for encoding annotation uncertainty into classification targets for machine learning tasks. SLEs encode annotations as Dirichlet distributions representing the subjective opinions of annotators using separate dimensions for first- and second-order uncertainty. Further, they reduce to the simpler representations of categorical and one-hot distributions in the absence of first- and second-order uncertainty, respectively. Learning SLEs using neural networks is also simple, requiring only a new output layer to compute the belief and uncertainty parameters, and the use of the KL-divergence objective function. 

\bibliography{sles}
\bibliographystyle{sles}

\appendix
\newpage
\appendix
\section{Notation}

\bgroup
\def\arraystretch{1.5}
\begin{tabular}{p{1.25in}p{3.25in}}
    $\displaystyle \fuse$           & Cumulative belief fusion operator \citep[12.5]{Josang_2016}. \\
    $\displaystyle \discount$       & Trust-discounting operator \citep[14.3]{Josang_2016}. \\
    $\displaystyle N$               & The number of training samples. \\
    $\displaystyle M$               & The number of annotators. \\
    $\displaystyle K$               & The number of class labels. \\
    $\displaystyle y^{(i)}_m$       & Label assigned to item $i$ by annotator $m$. \\
    $\displaystyle \omega^{(i)}_m$  & SLT opinion of annotator $m$ regarding the $i^{th}$ example. \\
    $\displaystyle \omega^{(i)}_{\lozenge}$   & Aggregated opinion over all annotators regarding the $i^{th}$ example. \\
    $\displaystyle \bm{b}^{(i)}_m$  & Belief vector for the $i^{th}$ example according to annotator $m$. \\
    $\displaystyle b^{(i)}_{m,k}$   & The belief value for the $k^{th}$ class label according to annotator $m$ for the $i^{th}$ example. \\
    $\displaystyle u^{(i)}_m$       & The subjective uncertainty of annotator $m$ regarding the $i^{th}$ example. \\
    $\displaystyle \bm{a}_m$        & The base rate or prior probabilities assigned over the $K$ events according to annotator $m$. \\
\end{tabular}
\egroup

\section{Experiment Details}
\label{app:hyperparams}

\begin{table}[h]
    \centering
    \begin{tabular}{lcccc|c}
        \toprule
              & \multicolumn{2}{c}{$c_m$} & \multicolumn{2}{c|}{$r_m$} & Annotation \\
              & $\alpha$ & $\beta$ &$\alpha'$ & $\beta'$       &    Uncertainty    \\
        \midrule
        Set 1 & 10       &    0    &    10    &    0           &  None             \\
        Set 2 & 10       &  1      &    10    &    1           &  Low              \\
        Set 3 & 10       &  10     &    10    &   10           &  Medium           \\
        Set 4 & 1        &  10     &    1     &   10           &  High             \\
        \bottomrule
    \end{tabular}
    \caption{Parameters for the annotation generative process given in \cref{sec:synth_data}.}
    \label{tab:data_params}
\end{table}

\section{Additional Results}
\label{app:add_results}

\newcommand{\mr}{\multirow}
\newcommand{\mc}{\multicolumn}
\begin{table}[h]
    \centering
    \begin{adjustbox}{center}
    \begin{tabular}{ll|ccc|ccc|ccc}
        \toprule
        \mc{2}{l|}{\mr{2}{*}{Annotations}} & \multicolumn{3}{c|}{(a)}         & \multicolumn{3}{c|}{(b)}        & \multicolumn{3}{c}{(c)}            \\
                               &            &  F1\up    &  JSD\dn  & NES\up    & F1\up    &  JSD\dn  & NES\up    &  F1\up   & JSD\dn   & NES\up       \\
        \midrule
        \mr{4}{*}{All}         & MV         &  0.601    &  0.515   &  0.923    & 0.783    & 0.437    & 0.923     &  0.187   & 0.645    & 0.923        \\
                               & Soft       &  0.601    &  0.312   &  0.955    & 0.784    & 0.301    & 0.948     &  0.187   & 0.463    & 0.928        \\
                               & CrowdTruth &  0.601    &  0.319   &  0.955    & 0.784    & 0.303    & 0.948     &  0.187   & 0.463    & 0.928        \\
                               & SLE        &{\bf0.804} &{\bf0.130}&{\bf0.989} &{\bf0.837}&{\bf0.135}&{\bf0.987} &{\bf0.611}&{\bf0.250}&{\bf0.973}    \\
        \midrule
        \mr{4}{*}{Filtered}    & MV         &  0.837    &  0.463   &  0.923    &{\bf0.929}& 0.427    & 0.923     &  0.802   & 0.477    & 0.923        \\
                               & Soft       &  0.837    &  0.176   &  0.978    &{\bf0.929}& 0.234    & 0.959     &  0.802   & 0.276    & 0.956        \\
                               & CrowdTruth &  0.837    &  0.192   &  0.978    &{\bf0.929}& 0.238    & 0.959     &  0.802   & 0.282    & 0.956        \\
                               & SLE        &{\bf0.872} &{\bf0.127}&{\bf0.990} & 0.923    &{\bf0.129}&{\bf0.987} &{\bf0.831}&{\bf0.212}&{\bf0.979}    \\
        \bottomrule
    \end{tabular}
    \end{adjustbox}
    \caption{Hard and soft metric results on the synthetic data corresponding to the plots in \cref{fig:synth_results}. (a) Average metrics over a range of reliabilities from very low to perfect, assuming perfect confidence. (b) Average metrics over a range of confidences from very low to perfect, assuming high reliability. (c) Average metrics over a range of confidences from very low to perfect, assuming low reliability. }
    \label{tab:synth_results_supp}
\end{table}

\end{document}